\documentclass[letterpaper, 10 pt, conference]{ieeeconf} \IEEEoverridecommandlockouts                            
\usepackage{graphicx}
\usepackage{eso-pic}

\IEEEoverridecommandlockouts                            
\usepackage{graphicx}
\usepackage{float}
\usepackage{atbegshi}
\usepackage{afterpage}
\usepackage{stfloats}
\usepackage{amsmath}
\usepackage{subfigure}
\usepackage{amssymb}
\usepackage{tabularx}
\usepackage{comment}

\usepackage{algorithm}
\usepackage{algpseudocode}
\usepackage{hyperref}
\usepackage{bm}
\usepackage[dvipsnames]{xcolor}
\usepackage{caption}

\usepackage{colortbl}

\usepackage[style=ieee]{biblatex}

\title{\LARGE \bf
SonarSweep: Fusing Sonar and Vision for Robust 3D Reconstruction via Plane Sweeping
}

\author{Lingpeng Chen$^{1}$, Jiakun Tang$^{1}$, Apple Pui-Yi Chui$^{2}$, Ziyang Hong$^{3*}$, and Junfeng Wu$^{1}$% <-this % stops a space
\thanks{$^{1}$Lingpeng Chen, Jiakun Tang, and Junfeng Wu are with the Chinese University of Hong Kong, Shenzhen. {\texttt{\{lingpengchen, jiakuntang, junfengwu\}@link.cuhk.edu.cn}}}% <-this % stops a space
\thanks{$^{2}$Apple Pui-Yi Chui is with the Chinese University of Hong Kong, Hong Kong \texttt{applepychui@cuhk.edu.hk}}% <-this % stops a space
\thanks{$^{3}$Ziyang Hong is with the Department of Automation, Harbin Institute of Technology, Shenzhen, China \texttt{{hongzy@hit.edu.cn}}}% <-this % stops a space
\thanks{$^{*}$Corresponding author: Ziyang Hong.}
}

\begin{document}

\maketitle

\begin{abstract}

Accurate 3D reconstruction in visually-degraded underwater environments remains a formidable challenge. Single-modality approaches are insufficient: vision-based methods fail due to poor visibility and geometric constraints, while sonar is crippled by inherent elevation ambiguity and low resolution. Consequently, prior fusion techniques rely on heuristics and flawed geometric assumptions, leading to significant artifacts and an inability to model complex scenes. In this paper, we introduce SonarSweep, a novel end-to-end deep learning framework that overcomes these limitations by adapting the principled plane sweep algorithm for cross-modal fusion between sonar and visual data. Extensive experiments in both high-fidelity simulation and real-world environments demonstrate that SonarSweep consistently generates dense and accurate depth maps, significantly outperforming state-of-the-art methods under challenging conditions, particularly in high turbidity. To foster further research, we publicly release our code and a novel dataset featuring synchronized stereo-camera and sonar data—the first of its kind—at \url{https://github.com/LIAS-CUHKSZ/SonarSweep}.

\end{abstract}

\section{Introduction}
    \label{Sec:Intro}
    Accurate 3D scene reconstruction is a fundamental capability for Autonomous Underwater Vehicles (AUVs), enabling critical applications like infrastructure inspection and environmental mapping~\cite{vanmiddlesworth2015mapping, nauert2023inspection}. These tasks demand operation in visually-degraded environments where turbid water and poor lighting pose formidable challenges to perception systems. 
    Achieving dense, metrically accurate, and geometrically coherent reconstructions under such conditions remains a significant and unsolved problem in robotics.
    % ~\cite{kaveti2025enhancing, huang2025visual}
    % ~\cite{ham2019computer}
    Vision-based methods, the standard for terrestrial 3D perception, are fundamentally unreliable underwater. First, the short baselines on compact AUVs render geometric triangulation ill-posed for objects beyond a few meters. Second, light scattering and absorption eradicate the high-frequency textures essential for robust stereo correspondence~\cite{angelopoulou2012limitations}. Active illumination techniques that project structured light can help~\cite{detry2018turbid}, but they fail in the highly turbid waters typical of real-world operations. Sonar offers robustness where vision fails, but its utility for 3D reconstruction is crippled by the inherent elevation ambiguity of 2D scans. Attempts to resolve this ambiguity in a single-modality system introduce critical flaws. Approaches that rely on vehicle motion to create multi-view observations~\cite{stereosonarwang, VerticalReconstruction} are critically dependent on accurate pose estimation—a fatal flaw in dynamic environments where odometry drift is unavoidable. Alternative solutions, such as using orthogonal sonars, are constrained by a severely limited overlapping field of view~\cite{OrthogonalSonar}.
    
    \begin{figure}[t]
        \centering
        \includegraphics[scale=0.26]{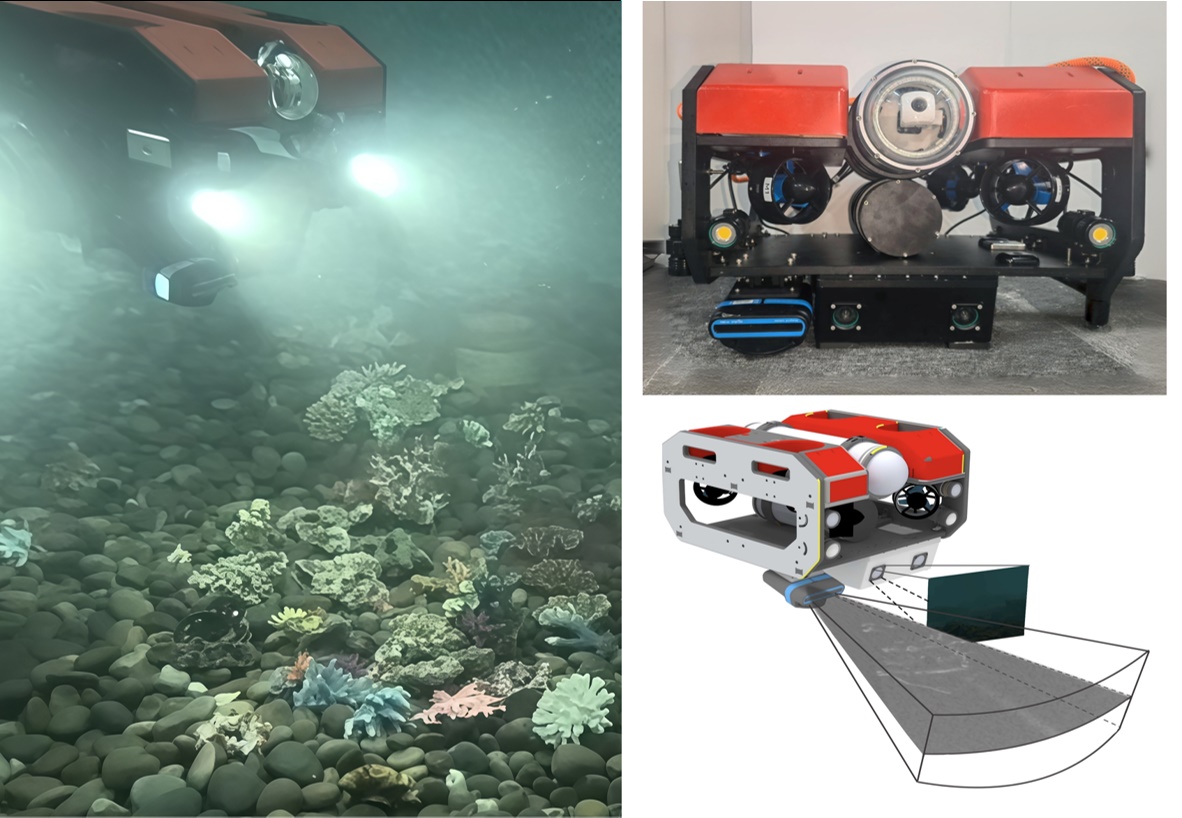}
        % \caption{The SonarSweep framework for robust opti-acoustic 3D reconstruction. (Left) Our experimental platform navigating a visually-degraded underwater environment. (Top Right) The sensor suite, integrating a stereo camera and a 2D imaging sonar. (Bottom Right) A conceptual diagram of our fusion strategy, which leverages dense camera imagery and sonar scans to achieve high-fidelity reconstruction.}
        \caption{\textbf{The SonarSweep System.} (Left) The experimental AUV in a challenging underwater environment. (Top Right) The integrated camera and sonar sensor suite. (Bottom Right) Conceptual diagram of the fusion approach.}
        \label{first_fig}
        \vspace{-15pt}
    \end{figure}

    % Given the complementary limitations of single-modality systems, fusing optical (high-detail) and acoustic (range-accuracy) data is a promising strategy. However, existing fusion techniques have failed to provide a complete solution. Vision-led SLAM systems that use sonar for scale correction fail when visual features are lost in turbid water~\cite{roznere2020underwater}.
    % Neural rendering frameworks such as AONeuS~\cite{AONeuS} and Z-Splat perform acoustic–optical fusion using NeRF or Gaussian Splatting representations. However, they require known sensor poses, are computationally expensive, and target volumetric reconstruction rather than direct depth estimation.
    % Real-time heuristic approaches rely on flawed geometric assumptions that introduce significant artifacts on non-vertical surfaces~\cite{OptiAcoustic}.
    % % Computationally expensive learning-based methods like AONeuS face challenges for real-time deployment~\cite{AONeuS}, while 

    Given the complementary limitations of single-modality systems, fusing optical (high-detail) and acoustic (range-accurate) sensing is a promising strategy. However, existing fusion techniques still fail to provide a complete solution. Vision-led SLAM systems that use sonar only for scale correction break down when visual features disappear in turbid water~\cite{roznere2020underwater}.
    Neural rendering frameworks such as AONeuS~\cite{AONeuS} and Z-Splat~\cite{qu2024z} perform acoustic–optical fusion using NeRF or Gaussian Splatting (GS) representations. However, these methods assume known sensor poses, are computationally expensive, and focus on volumetric reconstruction rather than direct depth estimation.
    Real-time heuristic approaches avoid heavy computation but rely on simplified geometric assumptions that introduce significant artifacts on non-vertical surfaces~\cite{OptiAcoustic}.
    
    To address this critical gap, we propose \textbf{SonarSweep}, a novel, end-to-end deep learning framework for dense and accurate underwater 3D reconstruction. Our method robustly fuses sonar and a monocular image by adapting the classic plane sweep algorithm to a learned, deep feature domain. We construct a multi-modal cost volume by differentiably warping sonar features into the camera’s reference frame across a set of depth hypotheses, allowing our framework to regress a dense and geometrically coherent depth map. Our contributions are threefold:
    \begin{itemize}
        \item The first adaptation of the deep plane sweep paradigm to the cross-modal fusion of sonar and visual data, overcoming the limitations of single-modality approaches.
        \item A comprehensive experimental validation showing that SonarSweep significantly outperforms state-of-the-art (SOTA) methods in challenging underwater conditions.
        \item The release of the first dataset of synchronized stereo-camera and imaging sonar data, along with our source code, to facilitate future research.
    \end{itemize}
            
\section{Related Works}
    \subsection{Opti-Acoustic Scene Reconstruction}
        % Fusing optical and acoustic sensors is a key strategy for robust 3D perception in turbid underwater environments where a single modality is insufficient. Heuristic methods, such as Opti-Acoustic~\cite{OptiAcoustic}, achieve real-time performance by circumventing direct cross-modal feature matching. The approach associates segmented image regions with clustered sonar returns and then back-projects a single depth value from the sonar to all pixels within the corresponding visual segment.
        
        % However, this efficiency relies on a critical geometric flaw: the assumption that all vertically aligned pixels share the same depth. This constrains the output to a series of "vertical curtains," introducing significant distortion on any inclined or complex surfaces. Consequently, while useful for mapping vertical structures, this methodology is fundamentally incapable of producing geometrically accurate models of general underwater scenes.

        Fusing optical and acoustic sensors is a key strategy for robust 3D perception in turbid underwater environments where a single modality is insufficient. Heuristic methods such as Opti-Acoustic~\cite{OptiAcoustic} achieve real-time performance by avoiding direct cross-modal feature matching, instead associating segmented image regions with clustered sonar returns and back-projecting a single sonar depth to all pixels within the corresponding visual segment.
        However, this efficiency relies on a critical geometric assumption: all vertically aligned pixels share the same depth. This constrains the reconstruction to a series of “vertical curtains,” introducing significant distortion on inclined or complex surfaces. 
        Recent neural rendering approaches, including AONeuS~\cite{AONeuS} and Z-Splat~\cite{qu2024z}, perform acoustic–optical fusion using NeRF or GS representations for 3D reconstruction. However, they assume known sensor poses and are not designed for direct pixel-wise depth estimation.
        Consequently, while useful for mapping vertical structures or improving volumetric reconstruction, these approaches remain unsuitable for accurate depth estimation in general underwater scenes.
    
    \subsection{Deep Plane Sweep Stereo}

        % The dominant paradigm for dense 3D reconstruction from multiple images is Deep Plane Sweep Stereo, which adapts a classic geometric algorithm into an end-to-end deep learning framework. Pioneering works like DPSNet~\cite{DPSNet} and MVSNet~\cite{MVSNet} established its core methodology. The process begins by extracting features from reference and source images using a neural network. The scene's depth is then discretized into a set of hypothesized, fronto-parallel planes. The key innovation is to project, or ``warp'', the source image features onto the reference camera's view for each depth plane.
        
        The dominant paradigm for dense multi-view 3D reconstruction is Deep Plane Sweep Stereo, which adapts a classical geometric algorithm into an end-to-end deep learning framework. Early works such as DPSNet~\cite{DPSNet} and MVSNet~\cite{MVSNet} established its core pipeline. The method first extracts features from reference and source images using a neural network, then discretizes scene depth into a set of hypothesized fronto-parallel planes. Source image features are subsequently projected (``warped'') onto the reference view for each depth plane. By comparing the reference features to the warped source features, a cost volume is constructed that encodes matching similarity for every pixel at every potential depth. A deep network then regularizes this volume, learning spatial and contextual relationships to refine the costs. From this regularized volume, a dense depth map is computed. This end-to-end pipeline replaces handcrafted similarity metrics with a powerful, learned correspondence model, achieving SOTA performance in vision-only reconstruction.

\section{Preliminaries and Notations}
    \label{sec:preliminaries}
    This section establishes the geometric models and notations for our system, which consists of a rigidly mounted and pre-calibrated pinhole camera and 2D forward-looking sonar (FLS).  We define a 3D point in the sonar coordinate system (Frame S) as \(\bm{P_s} = [X_s, Y_s, Z_s]^T\) and in the camera coordinate system (Frame C) as \(\bm{P_c} = [X_c, Y_c, Z_c]^T\). The transformation between these frames is defined by a rotation \(\bm{R}_s^c\) and translation \(\bm{t}_s^c\), such that \(\bm{P_c} = \bm{R}_s^c \bm{P_s} + \bm{t}_s^c\). The projection of \(\bm{P_c}\) onto the camera's image plane is denoted by the pixel coordinates \(\bm{p_c} = [u, v]^T\).
    
    Given a camera image \(\mathbf{I}_c\) and a corresponding 2D sonar scan \(\mathbf{S}_s\), our goal is to estimate a dense depth map \(\mathbf{D}_c\). This requires finding the depth value \(Z_c\) for every pixel \(\bm{p_c}\) by resolving the sonar's inherent geometric ambiguity for all points in the scene.
    
    \subsection{Forward-Looking Sonar Model}
        \label{subsec:SonarModel}
        \begin{figure}[htbp]
            \centering
            \includegraphics[width=0.6\columnwidth]{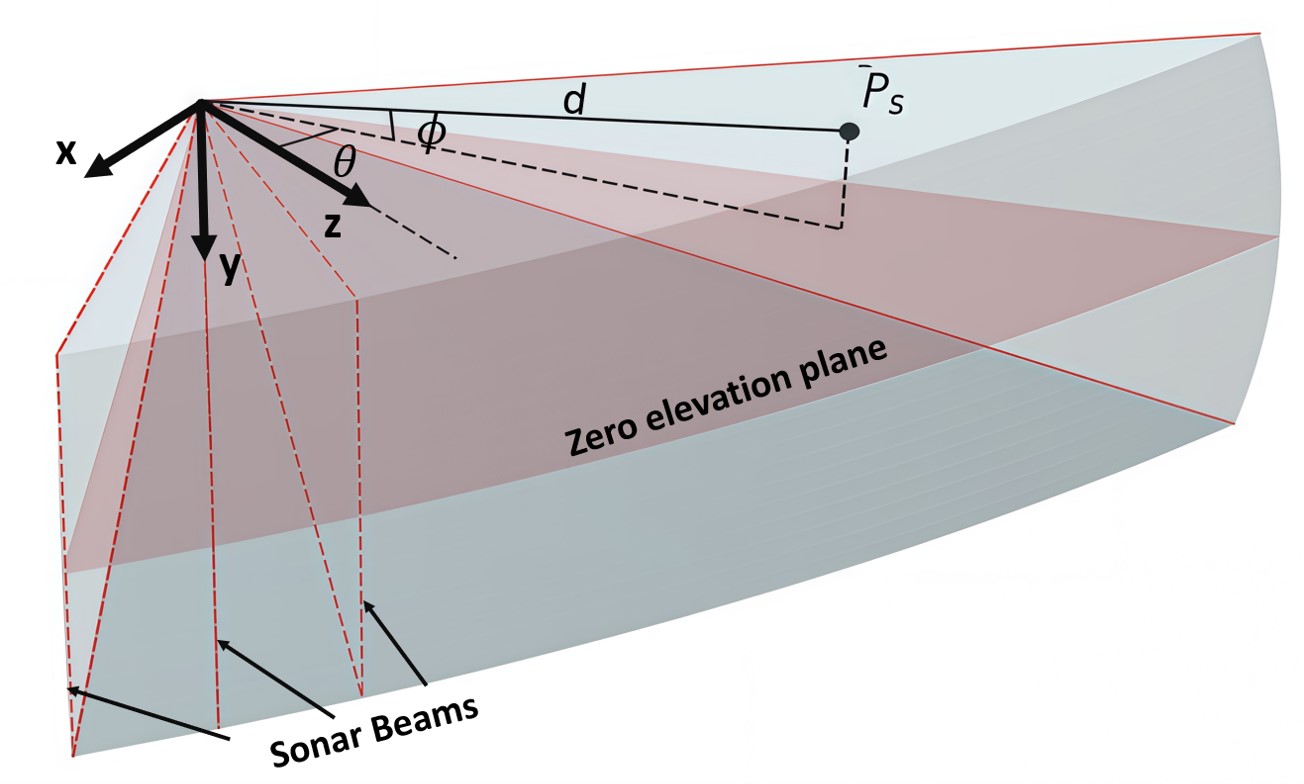}
            \caption{The Forward-Looking Sonar (FLS) sensor model. A 3D point \(\bm{P_s}\) is measured by its range \(d\) and bearing \(\theta\). The elevation angle \(\phi\) is collapsed during the projection, leading to ambiguity along a circular arc.}
            \label{fig:sonar_model}
            \vspace{-8pt}
        \end{figure}
        
        The primary challenge of using 2D FLS for 3D reconstruction is its inherent elevation ambiguity. A 3D point \(\bm{P_s}\) is described in spherical coordinates by its range \(d\), bearing \(\theta\), and elevation \(\phi\). As illustrated in Fig.~\ref{fig:sonar_model}, the sonar measures the range and bearing but collapses all elevation information, mapping the 3D point to a single 2D polar coordinate \(\bm{p_s} = (d, \theta)\). The relationship between spherical and Cartesian coordinates in the sonar frame is:
        \begin{equation}
            \bm{P_s} = \begin{bmatrix} X_s \\ Y_s \\ Z_s \end{bmatrix} = \begin{bmatrix} d \cos\phi \sin\theta \\ d \sin\phi \\ d \cos\phi \cos\theta \end{bmatrix}
            \label{eq:spherical_to_cartesian}
        \end{equation}
        
        Many FLS devices have a narrow vertical beamwidth, concentrating acoustic energy near the zero-elevation plane (\(\phi \approx 0\)). Following common practice~\cite{hurtos2015fourier}, we therefore adopt an \textbf{orthographic projection approximation} by assuming \(\cos\phi \approx 1\). This simplification effectively treats the 2D sonar image as a top-down view. Under this model, each sonar return \(\bm{p_s}\) constrains the horizontal position of a 3D point, while its elevation \(Y_s\) remains unresolved. Resolving this ambiguity by fusing the sonar data with the camera image is the core task of our work.
  
\section{Proposed Algorithm}
    \label{sec:proposed_algorithm}
    
\begin{figure*}[t]
        \centering
        \includegraphics[width=\textwidth]{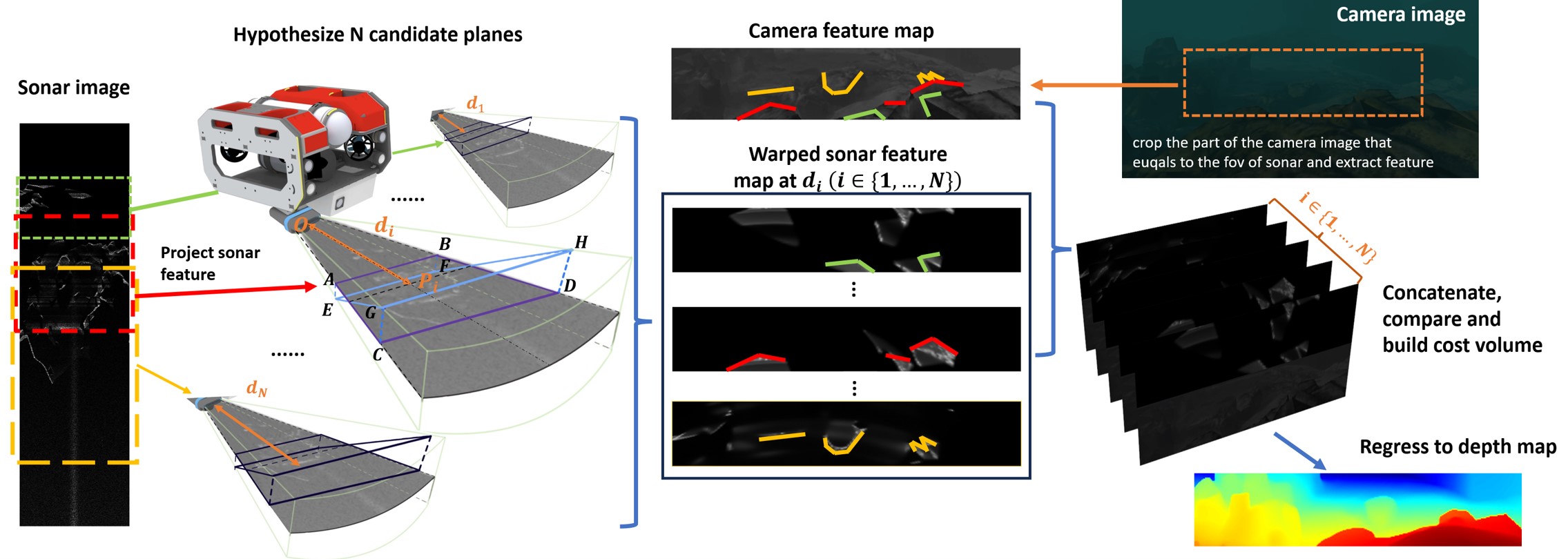}

        \caption{SonarSweep Pipeline. From a sonar and camera image pair, we extract feature maps using two encoders. The core of our method involves hypothesizing \(N\) candidate planes, onto which 2D sonar features are back-projected and differentiably warped into the camera's view. These warped feature maps are concatenated with the camera feature map to construct a multi-modal cost volume, which is regularized and regressed to a dense depth map. The colored lines in the central feature maps highlight the fundamental matching principle, where a structure finds its strongest correspondence only at the correct depth plane (\(d_i\)); the feature maps are illustrative visualizations of the high-dimensional vectors learned by the encoders.}
        \label{fig:flowchart}
        \vspace{-15pt}
    \end{figure*}

    The SonarSweep pipeline, illustrated in Fig.~\ref{fig:flowchart}, transforms the cross-modal reconstruction task into a structured, end-to-end learning problem. The framework consists of four main stages. First, two deep encoders extract multi-scale feature maps from the synchronized camera and sonar inputs respectively. For the camera, the image is cropped to the sonar's field of view and converted to grayscale; this encourages the network to learn robust, cross-modal \textbf{geometric similarities} rather than relying on spurious color-based correlations. Second, the 3D space is discretized into a set of \(N\) candidate planes, onto which 2D sonar features are back-projected and then differentiably warped into the camera's view. Third, a multi-modal cost volume is constructed by concatenating the camera feature map with the \(N\) warped sonar feature maps, encoding feature similarity across all hypothesized depths. Finally, a regularization network processes the cost volume, and a dense, metrically accurate depth map is regressed.
    
    \subsection{Sonar-Aligned Plane Hypothesization}
        \label{sec:hypothesize_and_sample}
        
        Instead of using the camera-centric, fronto-parallel planes common in stereo vision, our method discretizes the 3D space in alignment with the sonar's imaging geometry. From Fig.~\ref{fig:flowchart}, we hypothesize a set of \(N\) candidate planes within the sonar's field of view. As detailed in Fig.~\ref{fig:side_view}(a), each plane is parameterized by a fixed inclination angle \(\alpha\) and a unique distance \(\|OP_i\| = d_i\) from the sonar origin, for \(i \in \{1, ..., N\}\).
        
        This parameterization allows us to define a back-projection that maps any 2D sonar measurement onto a 3D point on each of these \(N\) planes. Leveraging the orthographic projection assumption (Section~\ref{subsec:SonarModel}), a sonar measurement \(\bm{p}_s = [d, \theta]^T\) is back-projected to its corresponding 3D point \(\bm{P}_s^i\) on the \(i\)-th plane as follows:
        \begin{subequations}
        \label{eq:backprojection}
        \begin{align}
            X_s^i &= d \sin(\theta) \\
            Y_s^i &= (d_i - Y_s^i) \tan(\alpha) = (d_i - d \cos(\theta)) \tan(\alpha) \\
            Z_s^i &= d \cos(\theta)
        \end{align}
        \end{subequations}
        % Here, \(X_s^i\) and \(Y_s^i\) are computed directly from the polar measurement, while the elevation \(Z_s^i\) is determined by the plane's geometry. This process effectively lifts the 2D sonar features into 3D, endowing them with the spatial information necessary for the subsequent warping stage.
        
        Here, \(X_s^i\) and \(Y_s^i\) are computed directly from the polar measurement, while the elevation \(Z_s^i\) is determined by the plane's geometry. This geometric transformation is the foundation of our learning approach; it is applied to rich feature maps extracted from parallel, multi-scale encoders for both the sonar and camera. This feature pyramid structure allows the network to learn robust correspondences by capturing both coarse and fine details. The process, therefore, effectively lifts the 2D sonar features into 3D, endowing them with the spatial information necessary for the subsequent warping stage.

    \subsection{Projective-Consistent Plane Sampling}
        % Our core objective is to maintain a constant geometric transformation between consecutive planes from the camera's perspective\footnote{This derivation assumes a zero-baseline configuration between the camera and sonar for illustrative clarity. The resulting geometric progression principle holds as a strong approximation for the practical, non-zero baseline case.}. This consistency requires that for two sonar points, when projected to camera frame, its pixel displacement from the projection of \(I^m\) (on plane \(i-1\)) to \(I^n\) (on plane \(i\)) equals the displacement from the projection of \(J^m\) (on plane \(i\)) to \(J^n\) (on plane \(i+1\)), as illustrated in Fig.~\ref{fig:side_view}(b). Assume \(I^m\) and \(J^m\) are on the ray \(m\), (i.e. have same projection point $m$),  \(I^n\) and \(J^n\) are on ray \(n\). This condition of projective consistency is met only if $\frac{\|OI^m\|}{\|OJ^m\|}=\frac{\|OI^n\|}{\|OJ^n\|} = \frac{\|OJ^m\|}{\|OK^m\|}$. Therefore, the ratio of the distances from the origin to consecutive planes must be constant:
        % \begin{equation}
        %     \frac{\|OP_i\|}{\|OP_{i-1}\|} = \frac{\|OP_{i+1}\|}{\|OP_i\|} = k
        % \end{equation}

        The strategic selection of the \(N\) plane distances is critical for effective feature matching. Standard methods that sample planes uniformly in depth or inverse-depth space~\cite{MVSNet} do not guarantee uniform pixel displacements for non-frontal scenes, which can impair the learned matching process. To address this, we introduce a \textbf{projective-consistent sampling} strategy designed to create uniform search steps in the camera's pixel space. 
        
        \begin{figure}[h]
            \centering
            \includegraphics[width=0.9\columnwidth]{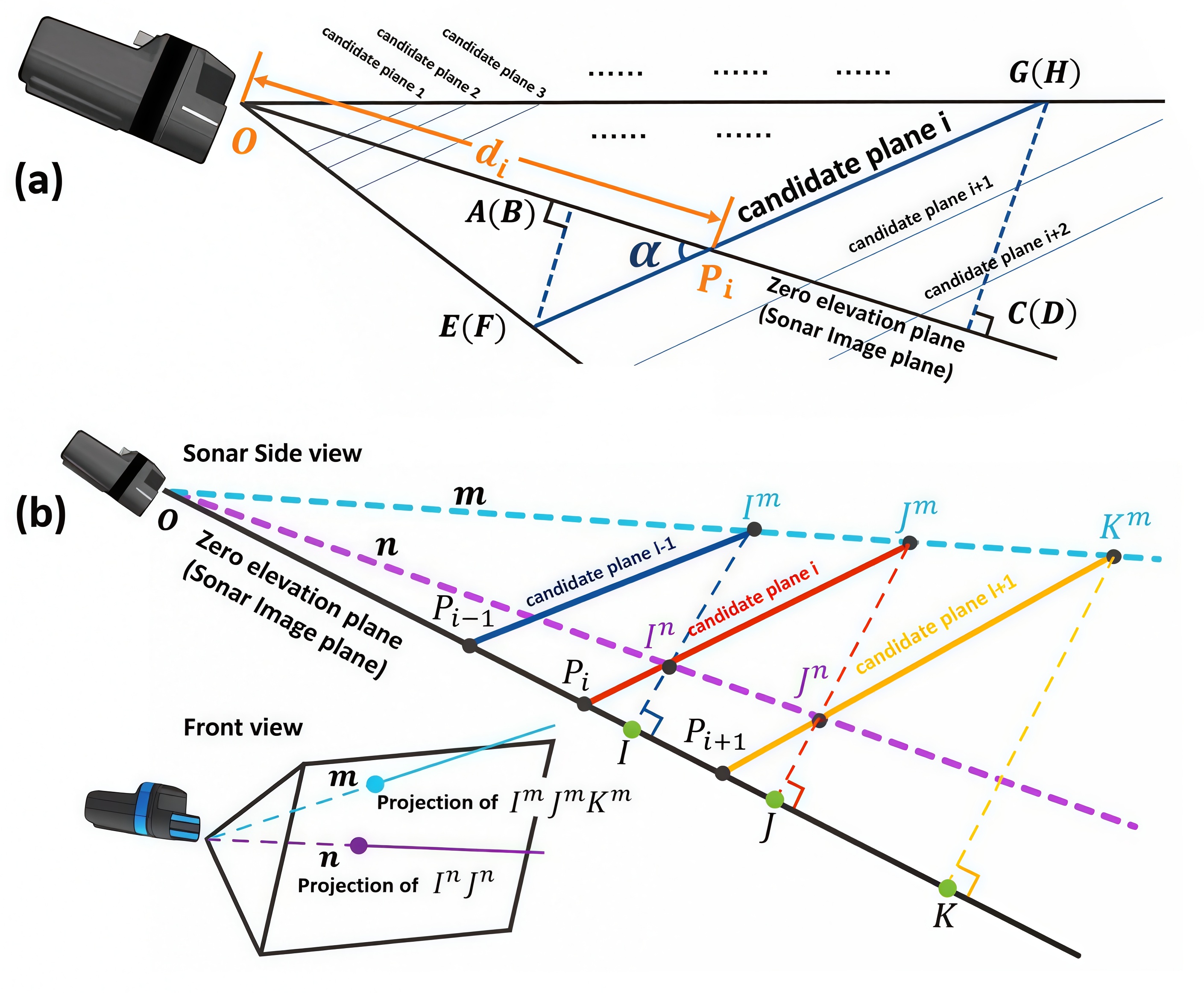}
            \caption{(a) Geometric parameterization of a candidate plane, defined by an inclination angle \(\alpha\) and a distance \(d_i\).
            \newline(b) Illustration of our projective-consistent sampling. \(I^m\) and \(I^n\) are 3D points back-projected from the 2D sonar measurement \(I\). This principle is applied consistently for all measurements (e.g., \(J\), \(K\)). To create uniform steps in the camera's pixel space, the hypothesized planes must be sampled in a geometric progression ($d_{i+1} = k \cdot d_i$).}
            \label{fig:side_view}
            \vspace{-10pt}
        \end{figure}
        
        Our core objective is to maintain a constant geometric transformation between consecutive planes from the camera's perspective\footnote{This derivation assumes a zero-baseline configuration between the camera and sonar for illustrative clarity. The resulting geometric progression principle holds as a strong approximation for the practical, non-zero baseline case.}. 
        As depicted in Fig.~\ref{fig:side_view}(b), this is achieved by enforcing projective consistency. 
        Consider the 3D points \(I^m\), \(J^m\) and \(K^m\), which lie on the same camera ray and thus project to the same image point \(m\). Likewise, \(I^n\) and \(J^n\) project to image point \(n\). 
        The consistency condition requires that the pixel displacement from \(m\) to \(n\) observed for the transition from plane \(i-1\) to \(i\) (via points \(I^m\) and \(I^n\)) is the same as for the transition from plane \(i\) to \(i+1\) (via points \(J^m\) and \(J^n\)).
        This is met only if the distances from the origin to these points form a geometric progression, i.e., $\frac{\|OI^m\|}{\|OJ^m\|} = \frac{\|OI^n\|}{\|OJ^n\|} = \frac{\|OJ^m\|}{\|OK^m\|}$.
        Therefore, the ratio of the distances to consecutive planes must be constant:
        \begin{equation}
            \frac{\|OP_i\|}{\|OP_{i-1}\|} = \frac{\|OP_{i+1}\|}{\|OP_i\|} = k
        \end{equation}
        where \(k\) is a constant scaling factor. This directly leads to our sampling formula, a geometric progression, where the distance to the \(i\)-th plane is given by:
        \begin{equation}
            d_i = d_0 \cdot k^{i-1}, \quad \text{for } i = 1, \dots, N
            \label{eq:geometric_explicit}
        \end{equation}
        where \(d_0 = d_{\text{min}}\). By creating uniform steps in this projective space, our strategy provides a more robust and stable basis for the learned feature matching. In our implementation, we set \(N=48\), \(d_0 = 0.5\)\,m, and \(k=1.05\), spanning the sonar's effective sensing range from 0.5\,m to approximately 5\,m.
    
    \subsection{Differentiable Warping via Ray-Plane Intersection}
        \label{sec:geo_transform}
        To construct the cost volume, we must warp the 3D sonar features, which reside on the hypothesized planes, into the camera's reference frame. This requires a differentiable mapping that, for each camera pixel \(\bm{p_c}\), finds its corresponding 3D location \(\bm{P_s^i}\) on the \(i\)-th candidate plane. Back-projecting a 2D pixel to a 3D point is inherently ill-posed, as the pixel's viewing ray contains infinite points. However, our plane hypotheses resolve this ambiguity by enforcing that the 3D point must lie on one of the known planes.

        We find this unique intersection point by formulating and solving a system of linear equations derived from two geometric constraints. This approach ensures the transformation is a closed-form, differentiable solution suitable for end-to-end learning.
        
        \subsubsection{The Planar Constraint}
        First, the point \(\bm{P_s^i}\) must lie on the \(i\)-th candidate plane. From Section~\ref{sec:hypothesize_and_sample}, this plane has a normal vector \(\bm{n_s} = [0, \cos(\alpha), \sin(\alpha)]^T\) and a known distance \(d_i\). This geometric fact yields our first linear equation for \(\bm{P_s^i}\):
        \begin{equation}
            \cos(\alpha)Y_s^i + \sin(\alpha)Z_s^i = d_i \sin(\alpha)
            \label{eq:plane_constraint}
        \end{equation}
        
        \subsubsection{The Camera Projection Constraint}
        Second, the point \(\bm{P_s^i}\) must project to the specified pixel coordinates \(\bm{p_c}=[u, v]^T\). This is described by the camera projection model: \(s [ u, v, 1 ]^T = \mathbf{K}_c (\mathbf{R}_s^c \bm{P_s^i} + \mathbf{t}_s^c)\), where \(s\) is the unknown depth in the camera frame. By defining \(\mathbf{M} = \mathbf{K}_c\mathbf{R}_s^c\) and \(\mathbf{C} = \mathbf{K}_c\mathbf{t}_s^c\), and letting \(\mathbf{m_j}\) be the j-th row of \(\mathbf{M}\), we can eliminate the unknown scalar \(s\) to obtain two additional linear equations for \(\bm{P_s^i}\):
        \begin{align}
            (u\mathbf{m_3} - \mathbf{m_1})^T \bm{P_s^i} &= C_1 - uC_3 \label{eq:linear_u} \\
            (v\mathbf{m_3} - \mathbf{m_2})^T \bm{P_s^i} &= C_2 - vC_3 \label{eq:linear_v}
        \end{align}
        
        \subsubsection{Solving the Linear System}
        Combining Equations~\eqref{eq:plane_constraint}, \eqref{eq:linear_u}, and \eqref{eq:linear_v} for each plane \(i\), we form a standard linear system \(\mathbf{A}_i\bm{P_s^i} = \mathbf{b}_i\):
        \begin{equation}
            \underbrace{
                \begin{bmatrix}
                    0 & \cos(\alpha) & \sin(\alpha) \\
                    \multicolumn{3}{c}{(u\mathbf{m_3} - \mathbf{m_1})^T} \\
                    \multicolumn{3}{c}{(v\mathbf{m_3} - \mathbf{m_2})^T}
                \end{bmatrix}
            }_{\mathbf{A}_i}
            \underbrace{
                \begin{bmatrix}
                    X_s^i \\ Y_s^i \\ Z_s^i
                \end{bmatrix}
            }_{\bm{P_s^i}}
            =
            \underbrace{
                \begin{bmatrix}
                    d_i \sin(\alpha) \\
                    C_1 - uC_3 \\
                    C_2 - vC_3
                \end{bmatrix}
            }_{\mathbf{b}_i}
            \label{eq:linear_system}
        \end{equation}
        This system has a unique solution, \(\bm{P_s^i} = \mathbf{A}_i^{-1}\mathbf{b}_i\), which can be computed efficiently. Performing this calculation for every pixel \(\bm{p_c}\) and every candidate plane \(i\) yields the complete set of 3D sampling coordinates. This grid is the crucial component that allows for the differentiable warping of sonar features into the camera's perspective.
    
    \subsection{Cost Volume Regularization and Depth Estimation}
        \label{sec:cost_volume_regression}
        With the differentiable warping grid established, the final stage of the pipeline transforms the geometric problem into a probabilistic estimation task, culminating in a dense depth map. This process involves constructing a multi-modal cost volume, regressing a probable depth for each pixel, and transforming this estimate into the final metric depth.

        \subsubsection{Cost Volume Construction and Regularization}
        
            The sampling grid \(\bm{P_s^i}\) computed in Section~\ref{sec:geo_transform} provides the precise coordinates to warp the sonar features into the camera's view. For each of the \(N\) candidate planes, we use this grid with a differentiable bilinear sampling mechanism to sample from the sonar feature map \(\mathcal{F}_s\). This generates \(N\) warped sonar feature maps, \(\mathcal{F}_s^i\), each aligned with the camera's perspective. A 4D cost volume \(\mathcal{C}\) of size \(H \times W \times N \times F\) is then constructed by concatenating the camera feature map \(\mathcal{F}_c\) with each of the \(N\) warped sonar feature maps. For a given pixel \((u, v)\), the feature vector at the \(i\)-th depth hypothesis is:
            \begin{equation}
                \mathcal{C}(u, v, i) = \text{Concat}(\mathcal{F}_c(u, v), \mathcal{F}_s^i(u, v))
            \end{equation}
            This volume, which encodes cross-modal similarity, is then processed by a 3D CNN. This network regularizes the costs by learning to enforce spatial and geometric consistency, producing a refined, single-channel cost volume \(\mathcal{C}'\) of size \(H \times W \times N\).
        
        \subsubsection{Differentiable Depth Regression}
            To obtain a continuous, sub-pixel accurate depth estimate from the discrete cost volume \(\mathcal{C}'\), we employ a soft-argmin operation. For each pixel, the matching costs across all \(N\) candidate depths are first converted into a probability distribution using the softmax function:
            \begin{equation}
                P(d_i | u, v) = \frac{\exp(-\mathcal{C}'(u, v, i))}{\sum_{j=1}^{N} \exp(-\mathcal{C}'(u, v, j))}
                \label{eq:softmax_prob}
            \end{equation}
            The final estimated plane distance \(\hat{d}(u, v)\) for the pixel is then computed as the expected value over this probability distribution:
            \begin{equation}
                \hat{d}(u, v) = \sum_{i=1}^{N} d_i \cdot P(d_i | u, v)
                \label{eq:expectation}
            \end{equation}
        
        \subsubsection{Final Metric Depth Transformation}
            The regression stage yields a per-pixel estimate of the most probable plane distance, \(\hat{d}(u,v)\), which exists in the sonar's geometric space. The final step is to transform this value into a dense, metric depth map in the camera's reference frame. For each pixel \(\bm{p_c}\), its corresponding 3D point \(\bm{\hat{P}}_c\) must simultaneously lie on its camera viewing ray and on the sonar plane defined by \(\hat{d}(u,v)\).
            
            By substituting the camera ray constraint (\(\bm{\hat{P}}_c = Z_c \cdot \mathbf{K}_c^{-1} [u, v, 1]^T\)) into the sonar plane constraint, we can solve directly for the unknown camera-frame depth, \(Z_c\). This yields a closed-form solution:
            \begin{equation}
                Z_c(u, v) = \frac{\hat{d}(u,v) \sin(\alpha) + (\mathbf{R}_s^c\bm{n_s})^T \mathbf{t}_s^c}{(\mathbf{R}_s^c\bm{n_s})^T (\mathbf{K}_c^{-1} [u, v, 1]^T)}
                \label{eq:zc_closed_form}
            \end{equation}
            Since \(Z_c\) represents the depth along the camera's principal axis, we convert it to a true Euclidean distance to generate the final metric depth map \(\mathcal{D}\):
            \begin{equation}
                \mathcal{D}(u, v) = |Z_c(u, v)| \cdot \left\| \mathbf{K}_c^{-1} \begin{bmatrix} u \\ v \\ 1 \end{bmatrix} \right\|_2
                \label{eq:final_depth}
            \end{equation}
            
        This entire pipeline, from feature extraction to the final depth computation, is fully differentiable, enabling end-to-end training of the SonarSweep network.
             
\section{Experiments and Results}
    \label{Sec:Experiment}
    This section validates our proposed camera-sonar fusion framework through a series of experiments. We analyze the model's performance in varied simulated and real-world conditions, compare it against SOTA baselines, and evaluate its robustness to environmental degradations such as turbidity.

    \begin{figure*}[tp]
        \centering
        \includegraphics[width=16.5cm]{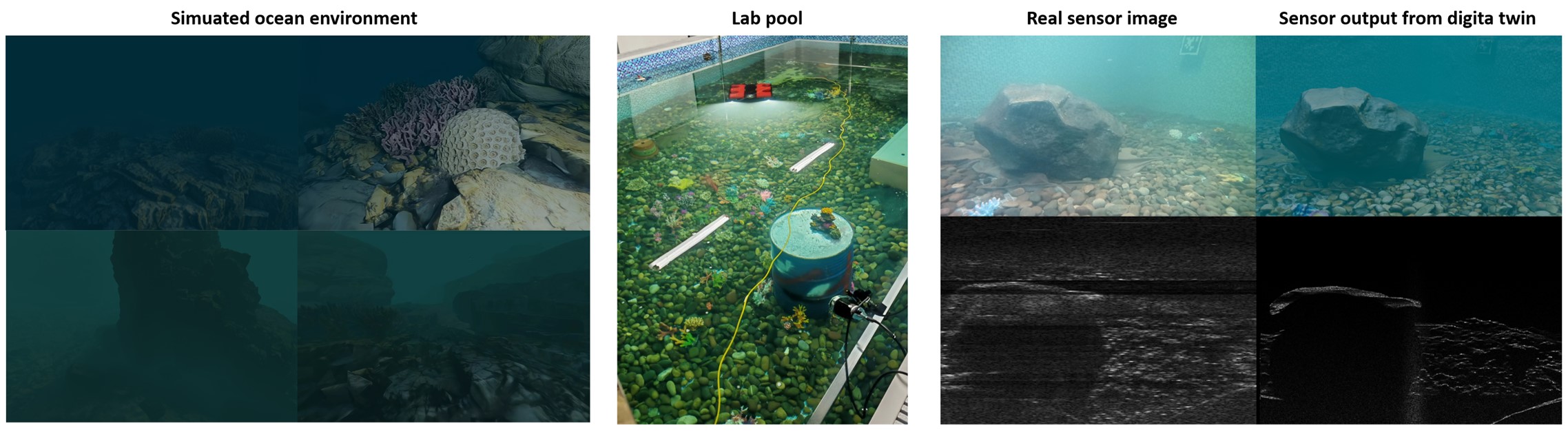}
        \vspace{-5pt}
        \caption{ From left to right: the simulated underwater world in OceanSim with varied water conditions; the physical lab pool setup; a real-world sensor suite; and the corresponding high-fidelity output from our digital twin.}
        \label{fig:experiment_environment}
        \vspace{-20pt}
    \end{figure*}
    
    \subsection{Experimental Setup}
        \subsubsection{Platform and Sensor Suite}
        Our experimental platform is a custom underwater vehicle equipped with a belly-mounted sensor pod, as shown in Fig.~\ref{first_fig}. The pod, which houses the primary perception sensors, is angled 15 degrees downwards to prioritize detailed seafloor mapping while maintaining forward-looking obstacle awareness. The sensor payload includes:
        \begin{itemize}
            \item \textbf{Stereo Camera:} A time-synchronized stereo camera with a 15\,cm baseline supports traditional stereo-vision algorithms, enabling fair comparison with ours. 
            \item \textbf{Imaging Sonar:} An Oculus M1200d forward-looking sonar operating in its high-frequency mode (2.1\,MHz) to maximize geometric detail. It provides a 60\textdegree{} horizontal and 12\textdegree{} vertical FOV within a 5m sensing range.
        \end{itemize}
        The extrinsic calibration parameters between the camera and sonar are directly derived from the robot's precise CAD design model and remain fixed in our algorithm.
        Model training was conducted offline on a desktop with an NVIDIA RTX 4090 GPU, while all inference and validation were performed exclusively on an NVIDIA Jetson Orin AGX. This setup mirrors the computational constraints of a field-deployable autonomous system.

        \subsubsection{Baselines for Comparison}
        To provide a comprehensive benchmark, we evaluate SonarSweep against three SOTA baselines, each representing a distinct sensing modality for dense depth estimation:
        \begin{itemize}
            \item \textbf{FoundationStereo~\cite{foundationstereo}:}  A SOTA vision-only baseline for stereo depth estimation. Renowned for its strong zero-shot generalization capabilities.
            \item \textbf{Multi-view Sonar Stereo~\cite{stereosonarwang}:} A learning-based, sonar-only method that adapts multi-view stereo principles to acoustic imagery.
            \item \textbf{Opti-Acoustic Fusion~\cite{OptiAcoustic}:} A representative geometry-based fusion method that offers a robust, real-time solution by matching visual segments with sonar returns.
        \end{itemize}
        These baselines were selected to ensure a fair evaluation against leading specialized methods in the vision-only, sonar-only, and heuristic fusion domains. We excluded methods like AONeuS \cite{AONeuS} and Z-Splat~\cite{qu2024z} from our comparison, as they are designed for neural rendering and are not suited for real-time and generalizable depth estimation.
        
    \subsection{Data Collection and Training Strategy}
        We adopted a dual-pronged data collection strategy, using a high-fidelity simulator for large-scale data generation and a physical testbed for real-world validation and fine-tuning.

        \subsubsection{Real-World Water Tank Data} Our physical experiments were conducted in a 4.5 × 10m water tank containing various objects like rocks, corals, and boulders to create a complex environment. Ground truth camera poses and depth maps were generated using the commercial photogrammetry software Agisoft Metashape~\cite{AgisoftMetashape}. This methodology is validated by its successful application in generating ground truth in the FLSea dataset~\cite{randall2023flsea}. The reconstruction quality was validated by an average RMS reprojection error of ~1.5 pixels, confirming its suitability as reliable ground truth. From this physical setup, we collected a total of 2,543 data points.

        % To augment the real-world dataset and directly study the simulation-to-reality (sim-to-real) gap, we also created a digital twin of our indoor tank within OceanSim. The right panel of Fig.~\ref{fig:experiment_environment} presents a qualitative comparison between the simulated and real environments at the same camera pose. While the simulated sonar image correctly captures the geometric shape of the rock, the real sonar image is corrupted by significant sensor noise and artifacts. This comparison highlights the substantial domain gap between the simulator and the real world.
        
        \subsubsection{Simulated Data Generation} We used the OceanSim~\cite{song2025oceansim} simulator, creating a digital twin of our robot that replicates its physical dimensions and sensor configurations. A large-scale underwater map was designed with diverse and complex terrain (e.g., rocks, corals, undulating seafloor). To increase data diversity, we rendered this environment under two distinct water conditions, simulating both clear inland and lower-visibility oceanic water, as shown in Fig.~\ref{fig:experiment_environment}. We collected 7686 synchronized data points from simulation, each including ground truth pose, a sonar image, a stereo image pair, and a dense depth map.

        % Our data preprocessing pipeline is tailored to standardize inputs and enhance salient features for both camera and sonar modalities.
            
        % For camera imagery, frames are first cropped to the sonar's field of view and converted to grayscale. This step isolates the relevant visual data and compels the network to learn robust, cross-modal geometric features rather than color-based correlations. Subsequently, Histogram Equalization is applied to enhance contrast.
        
        % For sonar data, we employ a preprocessing pipeline inspired by \cite{DSC} to mitigate ambient noise and remove static artifacts. As illustrated in Fig. \ref{fig:denoise_sonar}, the process begins by generating a static background model by averaging numerous frames captured without objects in the scene. Both the individual target frames and the averaged background image are then independently denoised using a pre-trained Swin-Conv-Unet (SCUNet) \cite{SCUNet}. Finally, the denoised background is subtracted from each denoised target frame. This procedure significantly improves the signal-to-noise ratio (SNR) by isolating the acoustic returns of the target object while removing static clutter and sensor-induced artifacts.

        \subsubsection{Data Preprocessing} 
\begin{figure}[h!]
            \centering
            % Replace with your actual image file
            \includegraphics[width=0.4\textwidth]{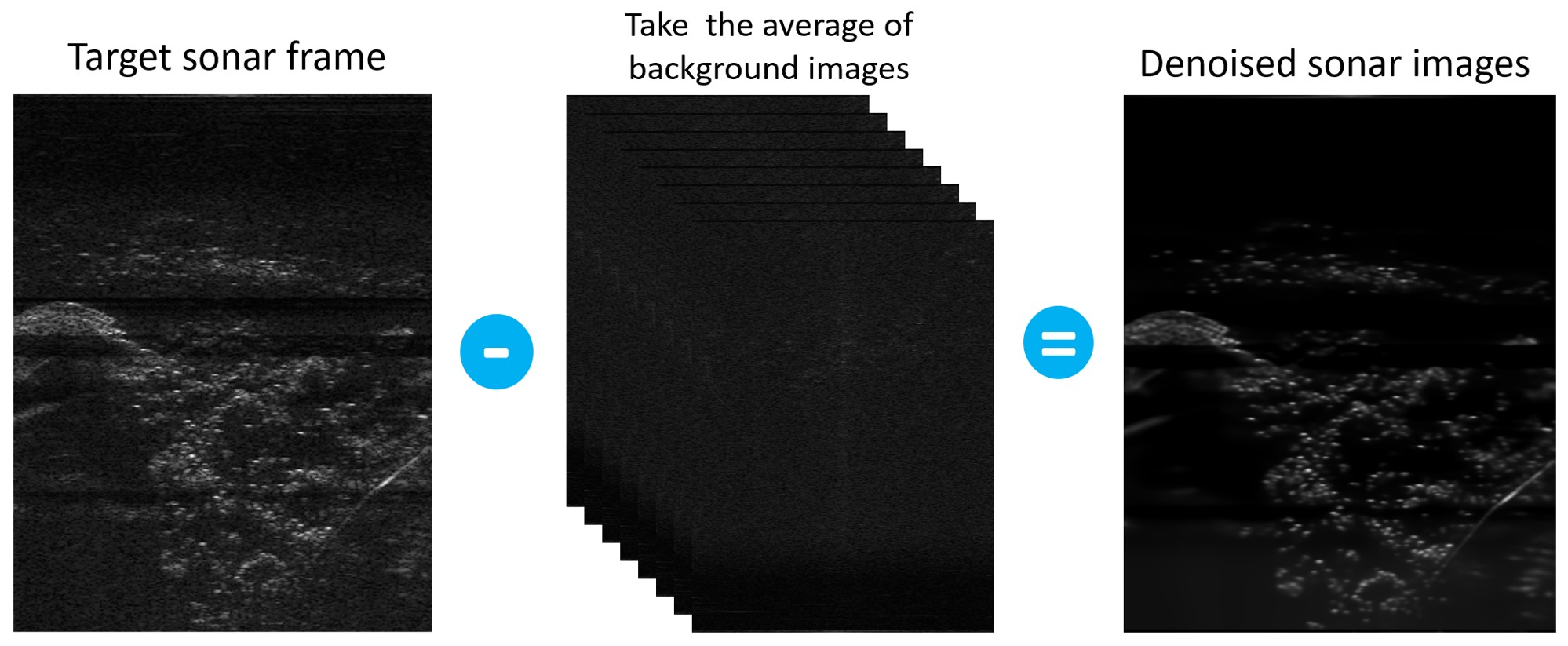}
            \caption{The sonar image preprocessing pipeline. }
            \label{fig:denoise_sonar}
            \vspace{-10pt}
        \end{figure}

        All data underwent a standardized preprocessing pipeline. Camera images were cropped to the sonar's FOV, converted to grayscale, and enhanced using histogram equalization to encourage the learning of geometric features. Sonar data was processed using a pipeline inspired by~\cite{DSC} to mitigate noise and artifacts, as shown in Fig.~\ref{fig:denoise_sonar}. First, a background model is created by averaging numerous frames captured without object in the scene. Next, both the individual target frames and this background model are independently denoised using a pre-trained Swin-Conv-Unet~\cite{SCUNet}. Finally, the denoised background is subtracted from each denoised target frame, significantly improving the signal-to-noise ratio by isolating the acoustic returns.
        \vspace{-10pt}
        % All data underwent a standardized preprocessing pipeline. Camera images were cropped to the sonar's FOV, converted to grayscale, and enhanced using histogram equalization to encourage the learning of robust geometric features. Sonar data was processed using a pipeline inspired by~\cite{DSC} to mitigate noise and artifacts, as shown in Fig.~\ref{fig:denoise_sonar}. This involved subtracting a denoised static background model from each denoised target frame, significantly improving the signal-to-noise ratio.

        \subsubsection{Sim-to-Real Training Strategy} To bridge the substantial domain gap between simulated and real-world data (see Fig.~\ref{fig:experiment_environment}, right panel), we employ a two-stage training strategy. The model is first pre-trained on the large-scale simulated dataset to learn generalizable cross-modal features. Subsequently, it is fine-tuned on the smaller, real-world dataset to adapt to the specific noise characteristics and sensor artifacts of the physical system.

    \subsection{Performance Analysis}

        We conducted a rigorous comparative analysis of SonarSweep against the three baselines across both simulated and real-world underwater environments. The evaluation demonstrates our method's superior performance in generating dense, accurate, and reliable depth maps.
        
        \begin{figure*}[t]
            \centering
            \includegraphics[width=16.5cm]{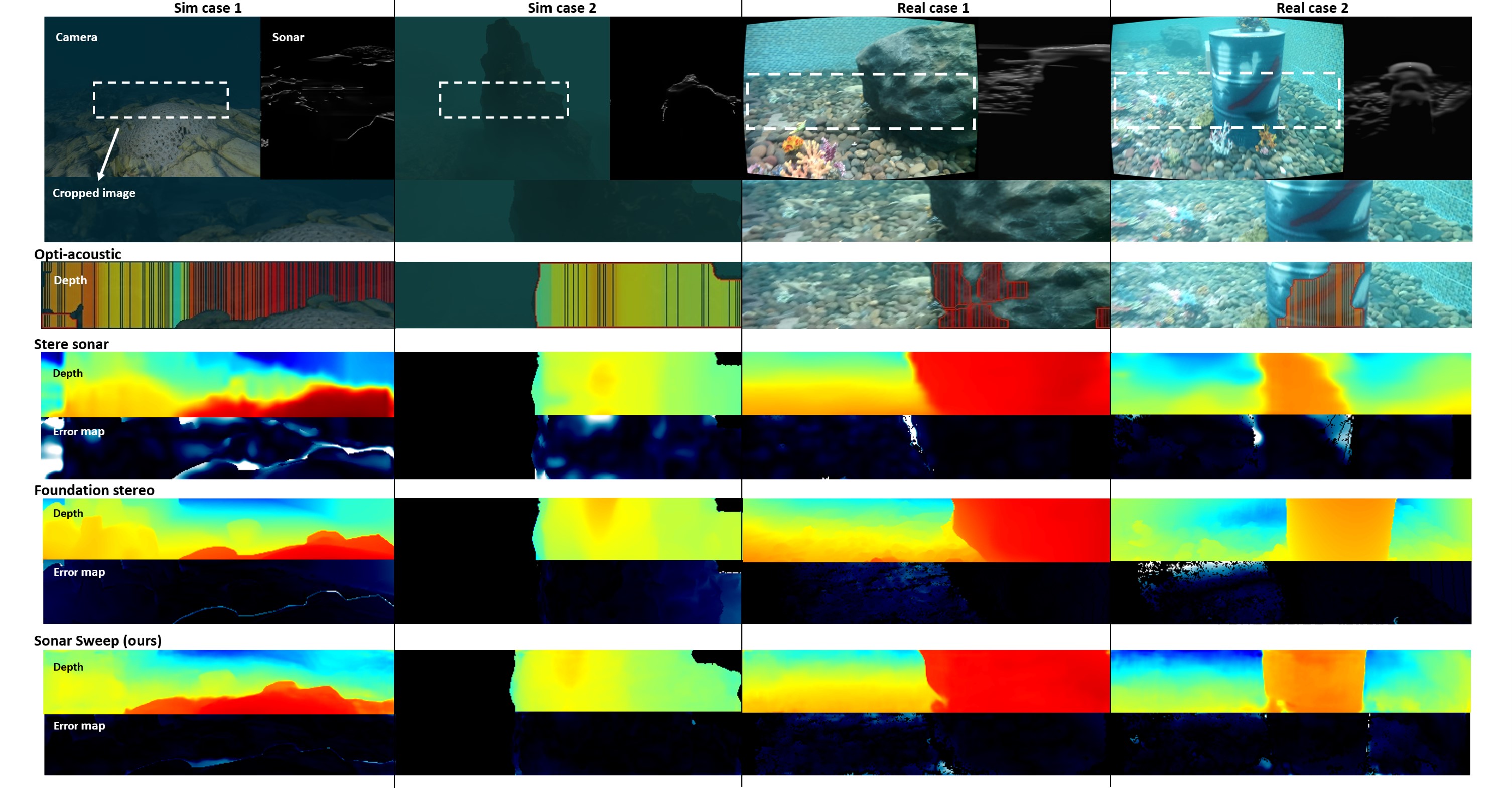}
            \caption{Qualitative comparison of each methods in simulated (Sim case 1 \& 2) and real-world (Real case 1 \& 2) scenarios. In the depth maps, warmer colors (red) are closer, while cooler colors (blue) are farther. In the error maps, brighter regions signify larger errors. Note that the output perspectives differ, as each algorithm uses a different native reference frame. }
            % Opti-acoustic and SonarSweep are aligned with the right camera, Foundation Stereo is aligned with the left camera, and Stereo Sonar is in the sonar coordinate system.}
            \label{fig:depth_visualization}
            \vspace{-10pt}
        \end{figure*}
        
        \subsubsection{Qualitative Comparison}
        Fig.~\ref{fig:depth_visualization} provides a qualitative comparison of the depth maps and error maps produced by each method. The results highlight the distinct failure modes of single-modality and heuristic fusion approaches. The \textbf{Opti-Acoustic} method produces sparse and often geometrically incorrect results, as its performance is limited by a fragile segmentation algorithm and a flawed ``vertical curtain'' assumption. It struggles to reconstruct continuous surfaces like the seafloor and fails to detect obstacles in the real-world cases.
        
        The sonar-only \textbf{Stereo Sonar} baseline is limited by the low resolution of acoustic imagery, resulting in depth maps with blurry edges and a lack of fine detail. Conversely, the vision-only \textbf{Foundation Stereo} excels at reconstructing nearby objects with sharp edges, but its accuracy rapidly degrades with distance due to the constrained stereo baseline. This leads to large-scale distortions on more distant surfaces.
        
        In contrast, \textbf{SonarSweep} consistently produces complete and geometrically accurate depth maps in all scenarios. By synergistically fusing the two modalities, our method leverages the camera's high-resolution texture for precise edge definition while relying on the sonar's robust range measurements to ensure accuracy over distance. This fusion overcomes the fundamental limitations of the individual sensors, resulting in a more reliable depth estimation.

        \begin{table}[htbp]
            \centering
            \caption{Quantitative Depth Estimation Performance}
            \label{tab:abs_rel}
            \begin{tabular*}{\linewidth}{@{\extracolsep{\fill}}lcccc}
            \hline \hline
            \multicolumn{5}{c}{\textbf{Simulation}} \\ \hline
                              & Abs Rel $\downarrow$ & Abs Diff $\downarrow$ & RMSE $\downarrow$  & a1 Acc. $\uparrow$ \\ \hline
            Foundation Stereo & 0.0509   & 0.1316    & 0.3807 & 0.9960 \\
            Stereo Sonar      & 0.0533   & 0.1387    & 0.2194 & 0.9622 \\
            \textbf{Ours}     & \textbf{0.0231}   & \textbf{0.0577}    & \textbf{0.0928} & \textbf{0.9951}  \\ \hline \hline
            \multicolumn{5}{c}{\textbf{Real-World}} \\ \hline
                              & Abs Rel $\downarrow$ & Abs Diff $\downarrow$ & RMSE $\downarrow$  & a1 Acc. $\uparrow$ \\ \hline
            Foundation Stereo & 0.0757   & 0.2437    & 0.3279 &  0.9506   \\
            Stereo Sonar      & 0.0691   & 0.1970    & 0.3131 & 0.9425 \\
            \textbf{Ours}     & \textbf{0.0382}   & \textbf{0.1064}    & \textbf{0.1479} & \textbf{0.9922}  \\ \hline
            \end{tabular*}
        \end{table}
        
        \begin{figure}[htbp]
            \centering
            \includegraphics[width=0.52\textwidth]{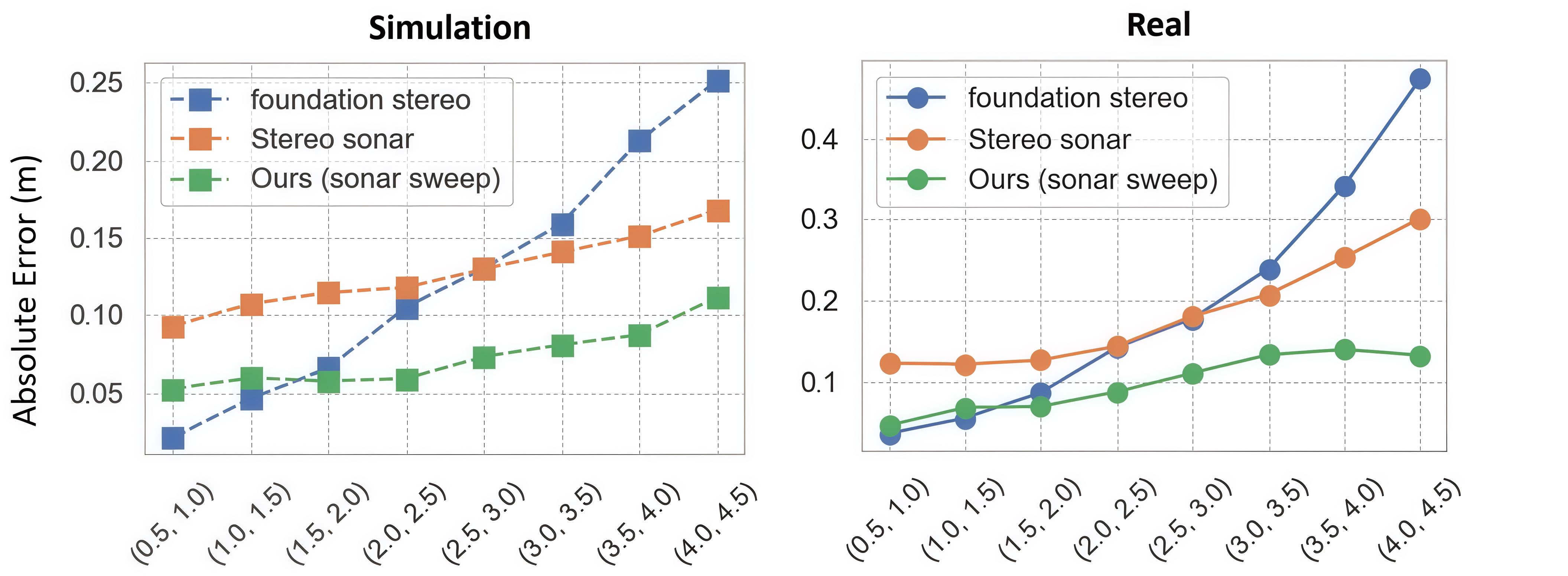}
            \caption{Quantitative comparison of absolute error versus distance in simulated (left) and real-world (right) datasets.}
            \label{fig:distance_error}
            \vspace{-15pt}
        \end{figure}
        
        \subsubsection{Quantitative Evaluation}
        The quantitative results, evaluated on 2,000 simulated and 1,000 real-world test sets, confirm the qualitative observations. 
        We evaluate depth quality using four standard metrics: Absolute Relative error (Abs Rel) and Absolute Difference (Abs Diff) for mean deviations, Root Mean Square Error (RMSE) for large-error penalties, and threshold accuracy ($a_1 < 1.25$).
        % We use four standard metrics: Absolute Relative error (Abs Rel), Absolute Difference (Abs Diff), Root Mean Square Error (RMSE), and threshold accuracy (a1).
        
        As shown in Tab.~\ref{tab:abs_rel}, our method significantly outperforms all baselines across all metrics on both datasets. The performance gap is particularly pronounced on the more challenging real-world data, highlighting our model's robustness to sensor noise and environmental variability.
        
        For a more granular analysis, Fig.~\ref{fig:distance_error} plots the estimation error as a function of depth. This reveals the complementary nature of our fusion approach. At close ranges ($<2m$), Foundation Stereo has highest accuracy, but its error grows sharply with distance. In contrast, the sonar-based methods exhibit more stable performance across the full range. SonarSweep effectively achieves the best of both worlds: it leverages high-resolution visual features to attain an accuracy comparable to Foundation Stereo at close ranges, while capitalizing on sonar's inherent precision to mitigate error growth over distance, maintaining a stability similar to the pure sonar method.

    \subsection{Robustness to Turbidity}
        A key advantage of opti-acoustic fusion is its potential for robust performance in degraded visual conditions. To validate this, we evaluated our model's resilience to turbidity by synthesizing poor visibility conditions on our real-world test data.
        
        \begin{figure*}[t]
            \centering
            \includegraphics[width=\textwidth]{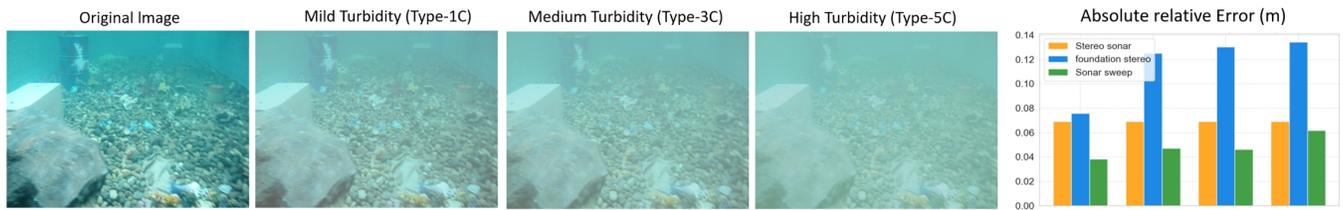}
            \vspace{-20pt}
            \caption{Synthesized images for mild, moderate, and high turbidity, followed by performance analysis}
            \label{fig:turbidity}
            \vspace{-15pt}
        \end{figure*}
        
        \subsubsection{Experimental Design}
        We synthesized turbidity on clear test tank images using the underwater Image Formation Model (IFM)~\cite{ahamed2022image}. Three conditions were simulated based on Jerlov coastal water types~\cite{jerlov1951photographic}: Type-1C (mild), Type-3C (moderate), and Type-5C (high turbidity). The turbid image \(I^c(x)\) was generated from the clear image \(J^c(x)\) using:
        \begin{equation}
            I^c(x) = J^c(x)(T_1^c)^d + (1 - (T_1^c)^d)B^c,
            \label{eq:combined_ifm}
        \end{equation}
        where \(c\) is the color channel, \(B^c\) is the ambient background light, \(d\) is object distance, and \(T_1^c\) is the spectral transmission rate. We used a representative distance of \(d=2.5\)\,m, with the specific transmission rates for each water type detailed in Tab.~\ref{tab:transmission_rates}.
        \vspace{-10pt}
        
        \begin{table}[htbp]
        \centering
        \caption{Spectral Transmission Rates ($T_1^c$) for Jerlov Types}
        \label{tab:transmission_rates}
        \begin{tabular}{|l|c|c|c|}
        \hline
        \textbf{Water Type} & \textbf{Red Channel} & \textbf{Green Channel} & \textbf{Blue Channel} \\ \hline
        Type-1C             & 0.75                 & 0.87                   & 0.88                  \\ \hline
        Type-3C             & 0.71                 & 0.80                   & 0.82                  \\ \hline
        Type-5C             & 0.67                 & 0.67                   & 0.73                  \\ \hline
        \end{tabular}
        \vspace{-10pt}
        \end{table}
        
        \subsubsection{Results}
        As illustrated in Fig.~\ref{fig:turbidity}, the performance of the Foundation Stereo degrades significantly as turbidity increases, as its feature matching relies entirely on visual clarity. In contrast, the acoustics-only Stereo Sonar is immune to this optical degradation, exhibiting a constant error rate across all conditions.       
        Our SonarSweep method consistently outperforms both baselines. While its error increases slightly in the most turbid scenario, it remains significantly more accurate than the individual modalities. This result confirms that our fusion strategy effectively leverages sonar data to compensate for the loss of visual information, ensuring reliable performance in challenging underwater environments.

        \section{Conclusion}
            \label{sec:conclusion}

            In this paper, we introduced \textbf{SonarSweep}, a novel, end-to-end deep learning framework that overcomes key challenges in underwater 3D reconstruction by fusing imaging sonar and camera data. Our core contribution is the successful adaptation of the deep plane sweep paradigm to this cross-modal problem. Extensive experiments demonstrated that SonarSweep significantly outperforms SOTA baselines, showing exceptional robustness at extended ranges and in turbid conditions. This work represents a significant step towards more reliable autonomous perception; to accelerate progress, we will release our source code and dataset, with future work aimed at integration into a full SLAM system for globally consistent mapping.

        \section*{Acknowledgment}
            This work was partly supported by the Guangdong Basic and Applied Basic Research Foundation under Grant No. 2024A1515240009, and the Shenzhen Science and Technology Program under Grant No. JCYJ20240813113609013.

\printbibliography{}

\end{document}